\documentclass[letterpaper]{article}
\usepackage{aaai17}

\usepackage{times}
\usepackage{helvet}
\usepackage{courier}
\usepackage{enumitem}
\usepackage{amssymb}
\usepackage{amsmath}
\usepackage{amsthm}
\usepackage{bbm}
\usepackage{graphicx}
\usepackage{subfigure}
\usepackage{hyperref}
\usepackage{multirow}
\usepackage[table,xcdraw]{xcolor}

\usepackage{tabularx, booktabs}

\makeatletter
\def\thm@space@setup{\thm@preskip=0pt
\thm@postskip=0pt}
\makeatother

\theoremstyle{plain}

\theoremstyle{definition}

\begin{document}

\title{Feasibility Study: Moving Non-Homogeneous Teams \\ in Congested Video
  Game Environments\thanks{Our demonstration paper resulted from the research
    of undergraduate student Jingxing Yang, primarily guided by graduate
    students Hang Ma and Liron Cohen. It was supported by NSF under grant
    numbers 1409987 and 1319966.}}  \author{ Hang Ma, Jingxing Yang, Liron
  Cohen, T. K. Satish Kumar \and Sven Koenig\\ Department of Computer Science,
  University of Southern
  California\\ \{hangma,jingxiny,lironcoh,skoenig\}@usc.edu,
  tkskwork@gmail.com }

\maketitle

\begin{abstract}
Multi-agent path finding (MAPF) is a well-studied problem in artificial
intelligence, where one needs to find collision-free paths for agents with
given start and goal locations. In video games, agents of different types
often form teams. In this paper, we demonstrate the usefulness of MAPF
algorithms from artificial intelligence for moving such non-homogeneous teams
in congested video game environments.

\end{abstract}

\section{Introduction}

Path finding is a component of many video games. For example, agents in
turn-based or real-time strategy games need to plan collision-free paths from
their current locations to their goal locations, often in dynamic and
congested environments. Moving the agents in a team rather than individually
makes it easier for players to control hundreds of agents.  Furthermore, the
agents often have types and thus form a non-homogeneous team. In Dragon Age:
Origins, for example, a player moves teams of agents, where a team might
consist of mages, warriors, and rogues. Agents of the same type form a group
within the team because they can interchange their goal locations. In our
example, the mages thus form the first group, the warriors form the second
group, and the rogues form the third group. Each goal location reserved for a
mage, warrior, or rogue can be occupied only by a mage, warrior, or rogue,
respectively, but it does not matter by which one. A similar situation arises
in Age of Empires for archers, spearmen, and knights.

We therefore study the problem where a player moves a non-homogeneous team by
occasionally specifying goal locations for them, for example, after observing
new parts of the environment. The player may specify new goal locations even
before all agents have reached the previously specified goal locations. We
demonstrate the usefulness of multi-agent path finding (MAPF) algorithms from
artificial intelligence for finding collision-free paths for all agents.

\begin{figure*}[t]
  \centering
  \includegraphics[width=0.32\textwidth]{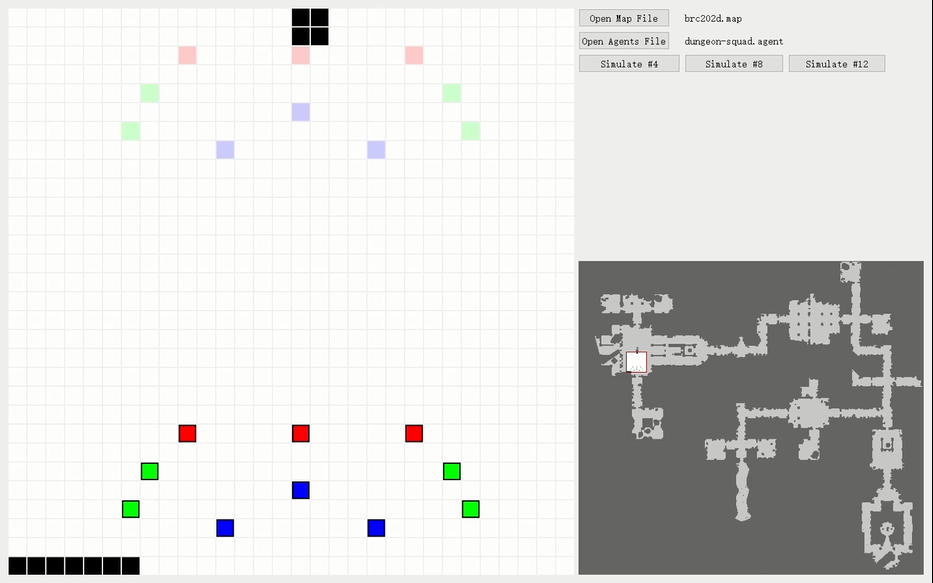}~\includegraphics[width=0.32\textwidth]{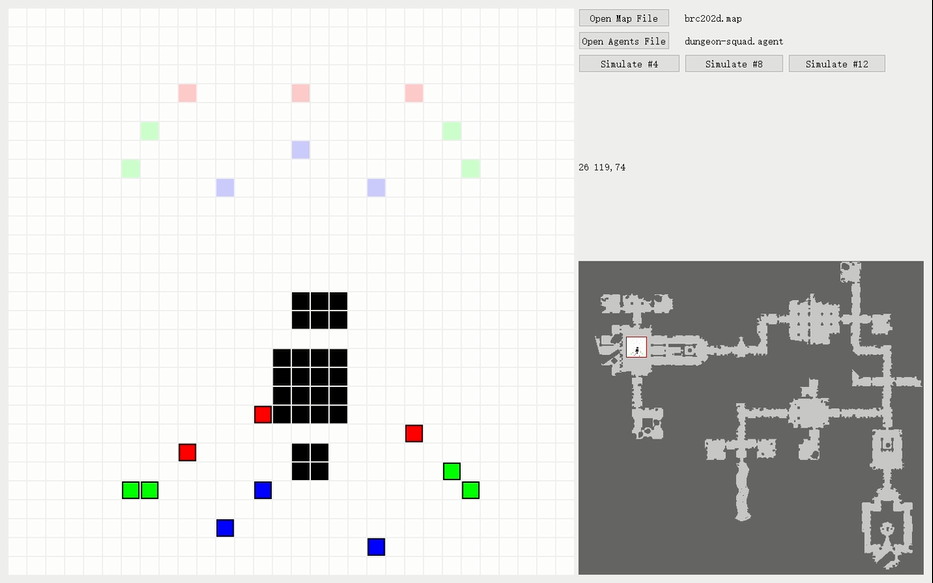}~\includegraphics[width=0.32\textwidth]{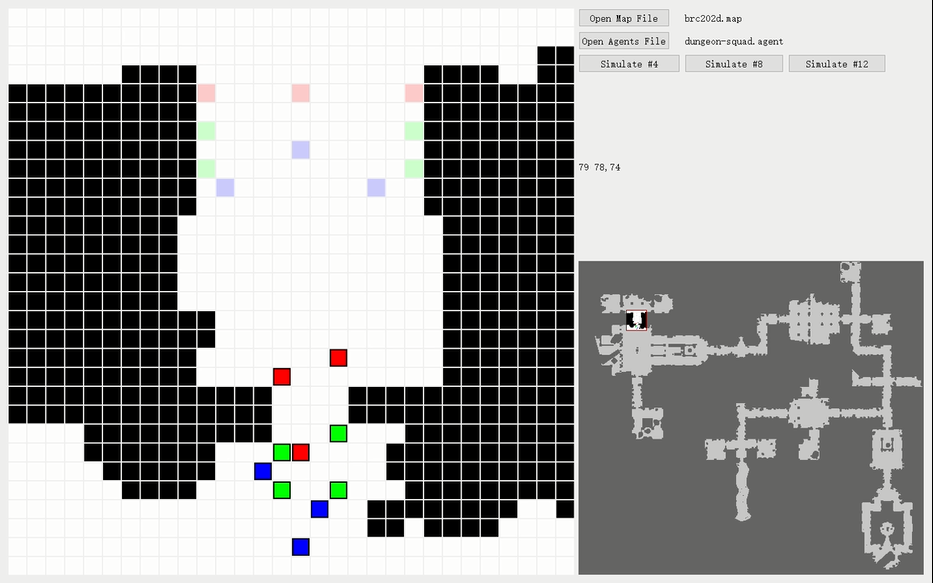}\\
  \includegraphics[width=0.32\textwidth]{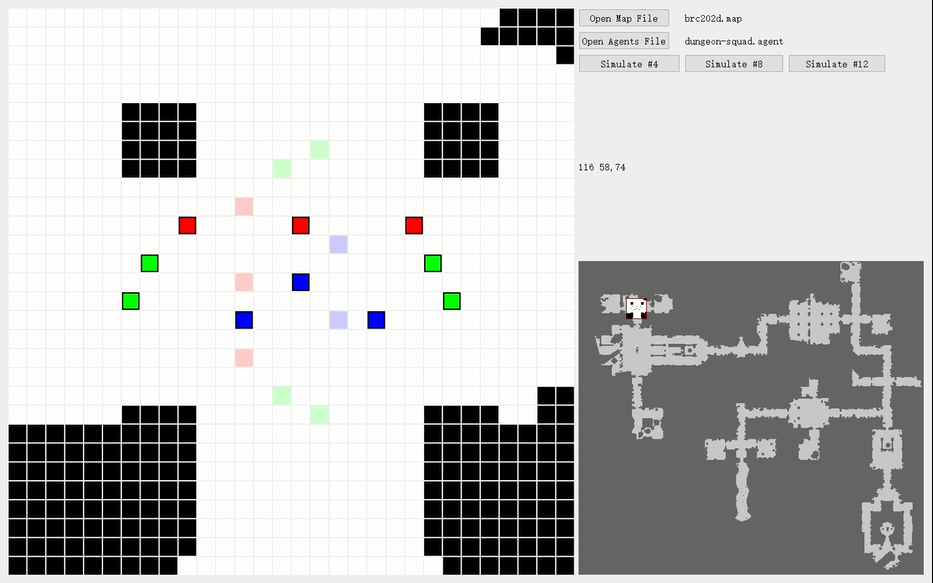}~\includegraphics[width=0.32\textwidth]{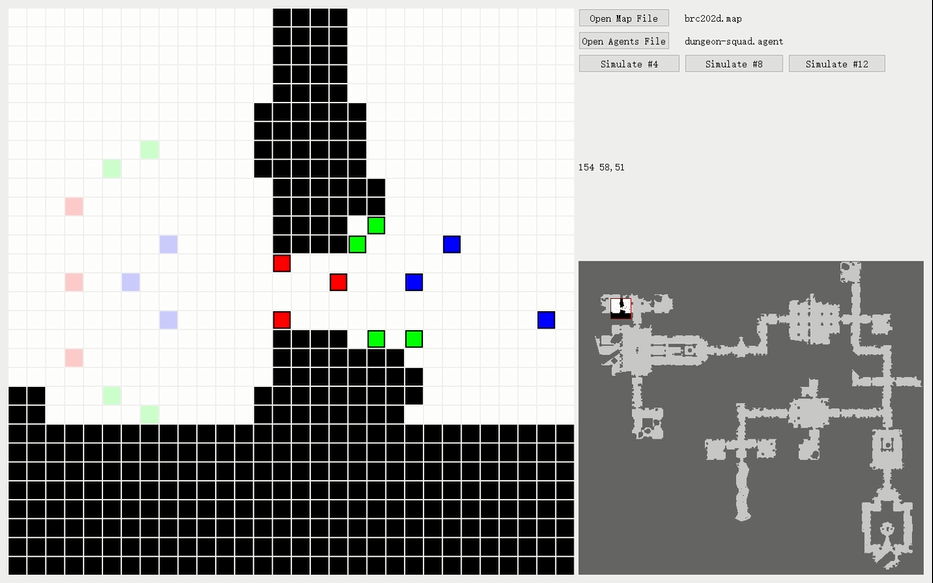}~\includegraphics[width=0.32\textwidth]{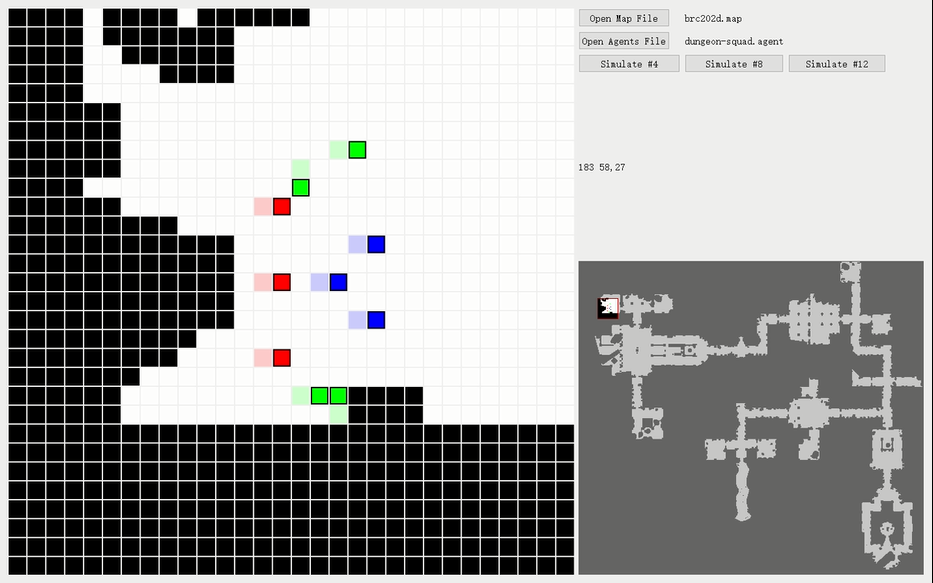}
  \caption{Screenshots of moving ten agents in three groups with a wide goal
    pattern. New goal locations are specified every twelve time steps. From
    top-left to bottom-right: (1) agents start; (2) agents avoid obstacles;
    (3) agents pass through a narrow passageway; (4) agents turn (since the
    goal pattern was rotated by ninety degrees); (5) agents pass through
    another narrow passageway; and (6) agents reach their final goal
    locations.}\label{fig:demo}
\end{figure*}

\section{Multi-Agent Path Finding}

MAPF is NP-hard to solve optimally \cite{YuLav13AAAI,MaAAAI16}. It can be
solved via reductions to other well-studied combinatorial problems
\cite{Surynek15,YuLav13ICRA,erdem2013general} or by dedicated optimal,
bounded-suboptimal, or suboptimal MAPF algorithms
\cite{ODA11,EPEJAIR,DBLP:journals/ai/SharonSGF13,wagner15,DBLP:journals/ai/SharonSFS15,ICBS,CohenUK16,WHCA,WHCA06,PushAndSwap,PushAndRotate,WangB11}. Many
MAPF algorithms have been used on maps from video games \cite{WHCA}. See
\cite{MaWOMPF16,FelnerSOCS17} for longer surveys on MAPF algorithms.

MAPF has recently been generalized in different directions
\cite{HoenigICAPS16,MaWOMPF16,HoenigIROS16,MaAAAI17,MaAAMAS16,MaAAMAS17}.
Target Assignment and Path Finding (TAPF) is a variant of MAPF that allows
agents in the same group to interchange their goal locations \cite{MaAAMAS16}
and thus applies to non-homogeneous teams. During execution, one can maintain
user-specified safety distances between agents and adhere to their kinematic
constraints \cite{HoenigICAPS16,HoenigIROS16}.

We use Conflict-Based Min-Cost-Flow (CBM) \cite{MaAAMAS16}, a state-of-the-art
optimal TAPF algorithm. CBM is a two-level algorithm that minimizes the
makespan (that is, the earliest time when all agents reach their goal
locations). On the upper level, CBM performs a best-first search on a
collision tree and resolves collisions between agents in different
groups. Each high-level node contains a set of constraints and a path for each
agent that obeys these constraints. On the lower level, CBM uses a
polynomial-time min-cost max-flow algorithm \cite{Successive} on a
time-expanded network.

Each time the player specifies new goal locations, CBM is called to solve a
new TAPF instance from the current locations of the agents to their newly
specified goal locations. We use cost one for move actions and cost zero for
wait actions to avoid unnecessary move actions, such as moving forward and
then immediately backward. We also use a large cost for actions that result in
collisions between agents in different groups to make CBM more efficient, as
discussed in \cite{MaAAMAS16}.

\section{Demonstration}

We use CBM with ten agents in three groups in the video game environment {\tt
  brc202d} \cite{sturtevant2012benchmarks} from Dragon Age: Origins on a
2.5GHz Intel Core i5-2450M with 4GB of RAM. CBM plans in a window of size 30
cells by 30 cells around the agents. The window typically contains many rows
in the direction of the next goal locations and few rows in the opposite
direction.  The goal locations are specified manually. They are typically
about twenty cells away from the current locations of the agents. We use two
goal patterns (that is, layouts of the goal locations), namely (nineteen
cells) wide and (seven cells) narrow, and three update frequencies of the goal
locations, namely every four, eight, and twelve time steps. Our videos are
available at \url{http://idm-lab.org/bib/abstracts/Koen17k.html}.

\begin{table}  
\small
\centering
\caption{Experimental results.}\label{tab:results}
\begin{tabular}{|c|ccc|}
  \hline
  instance & makespan & CBM calls & \begin{tabular}[c]{@{}c@{}}average\\running time (s)\end{tabular} \\
  \hline
  wide-4  & 193 & 50 & 0.146 \\
  wide-8  & 190 & 25 & 0.202 \\
  wide-12 & 184 & 17 & 0.153 \\
  narrow-4  & 158 & 40 & 0.139 \\
  narrow-8  & 157 & 20 & 0.139 \\
  narrow-12 & 156 & 14 & 0.151 \\
  \hline
\end{tabular}
\end{table}

Table \ref{tab:results} reports the makespan, the number of calls to CBM, and
the average running time per call (in seconds) for each combination of width
of the goal pattern and update frequency of the goal locations. Figure
\ref{fig:demo} shows screenshots for a wide goal pattern and an update
frequency of twelve time steps (called wide-12 in the table). The first group,
shown in red, consists of three warriors. The second group, shown in green,
consists of four rogues.  Finally, the third group, shown in blue, consists of
three mages. The goal pattern is shown in a lighter shade of the same color as
the agents of the same group. CBM is called every twelve time steps to solve a
new TAPF instance from the current locations of the agents to their newly
specified goal locations.

\section{Conclusions and Future Work}

Our experimental results show that CBM runs sufficiently fast to find
collision-free paths for small numbers of agents over short distances in
real-time, even in congested video game environments. We are currently working
on a system that moves agents in formation over longer distances.  Formations,
for example, keep the agents safe by making the warriors move in front of the
formation, the rogues on both sides, and the mages in the back. However,
formations often have to be compromised temporarily in congested video game
environments, for example, when agents move through passageways that are
narrower than their formation. In our demonstration, the goal patterns
correspond to the above formation, yet the agents do not always restore the
formation during execution immediately when possible, which is an issue that
our system will address. Our system will use swarm-based approaches in simple
parts of the video game environment (because they run fast) and switch to CBM
in congested parts (because swarm-based approaches fail in them), where it
automatically determines appropriate intermediate goal locations that trade
off between a small running time, a small makespan (that is, short paths) and
a close adherence to the player-specified formation (where the tradeoff can be
specified by the player as well).

\newpage
%\small
\bibliographystyle{aaai}
\bibliography{references}
\end{document}